\documentclass[sigconf]{acmart}
\renewcommand\footnotetextcopyrightpermission[1]{} % removes footnote with conference information in first column
\usepackage{nopageno}
\usepackage{amsmath}

\AtBeginDocument{%
  \providecommand\BibTeX{{%
    \normalfont B\kern-0.5em{\scshape i\kern-0.25em b}\kern-0.8em\TeX}}}

\acmConference[WSDM '22]{WSDM '22: ACM International on Web Search and Data Mining}{Feb 21--25, 2022}{Phoenix, AZ}
\acmBooktitle{ACM International on Web Search and Data Mining,
  Feb 21--25, 2022, Phoenix, AZ}
\acmPrice{15.00}
\acmISBN{978-1-4503-XXXX-X/18/06}

\usepackage{algorithm}

\usepackage{algorithmic}
\usepackage{xcolor}
\usepackage{booktabs}
\usepackage{multirow}

\usepackage{bm}

\begin{document}

\title{Graph Few-shot Class-incremental Learning}

%
% The "author" command and its associated commands are used to define
% the authors and their affiliations.
% Of note is the shared affiliation of the first two authors, and the
% "authornote" and "authornotemark" commands
% used to denote shared contribution to the research.
\author{Zhen Tan}
\affiliation{%
  \institution{Arizona State University}
%   \city{Tempe}
  \country{}
%   \postcode{85281}
}
\email{ztan36@asu.edu}

\author{Kaize Ding}
\affiliation{%
  \institution{Arizona State University}
%   \city{Tempe}
  \country{}
%   \postcode{85281}
}
\email{kaize.ding@asu.edu}

\author{Ruocheng Guo}
\affiliation{%
  \institution{City University of Hong Kong}
%   \streetaddress{}
%   \city{Kowloon}
  \country{}
  }
\email{ruocheng.guo@cityu.edu.hk}

\author{Huan Liu}
\affiliation{%
  \institution{Arizona State University}
%   \city{Tempe}
  \country{}
%   \postcode{85281}
}
\email{huanliu@asu.edu}

%
% By default, the full list of authors will be used in the page
% headers. Often, this list is too long, and will overlap
% other information printed in the page headers. This command allows
% the author to define a more concise list
% of authors' names for this purpose.
\renewcommand{\shortauthors}{}
\renewcommand{\algorithmiccomment}[1]{#1}

\begin{abstract}
The ability to incrementally learn new classes is vital to all real-world artificial intelligence systems. A large portion of high-impact applications like social media, recommendation systems, E-commerce platforms, etc. can be represented by graph models. In this paper, we investigate the challenging yet practical problem, \textit{Graph Few-shot Class-incremental} (Graph FCL) problem, where the graph model is tasked to classify both newly encountered classes and previously learned classes. Towards that purpose, we put forward a Graph Pseudo Incremental Learning paradigm by sampling tasks recurrently from the base classes, so as to produce an arbitrary number of training episodes for our model to practice the incremental learning skill. Furthermore, we design a Hierarchical-Attention-based Graph Meta-learning framework, HAG-Meta. We present a task-sensitive regularizer calculated from task-level attention and node class prototypes to mitigate overfitting onto either novel or base classes. To employ the topological knowledge, we add a node-level attention module to adjust the prototype representation. Our model not only achieves greater stability of old knowledge consolidation, but also acquires advantageous adaptability to new knowledge with very limited data samples. Extensive experiments on three real-world datasets, including Amazon-clothing, Reddit, and DBLP, show that our framework demonstrates remarkable advantages in comparison with the baseline and other related state-of-the-art methods.  
\end{abstract}


\vspace{-0.3cm}

\maketitle

\section{Introduction}
Graph-structured data, such as citation graphs \cite{tang2008arnetminer}, biomedical graphs \cite{subramanian2005gene}, and social networks \cite{qi2011exploring, hamilton2018inductive}, are nowadays ubiquitous in different real-world applications. Recently, a spectrum of Graph Neural Networks (GNNs) \cite{cao2016deep, defferrard2016convolutional, henaff2015deep, kipf2016semi, hamilton2018inductive, velivckovic2017graph, xu2018powerful} has been proposed to model graph-structured data by transforming node features and propagating the embedded features along the graph structure. As a central task in graph machine learning, semi-supervised node classification aims to infer the missing labels for unlabeled nodes. By capturing the information carried by both labeled and unlabeled nodes as well as the relations between them, GNNs are able to achieve superior performance over other approaches.

However, most of the existing work on node classification primarily focuses on a single task, where the model is tasked to classify the unlabeled nodes to a fixed set of classes \cite{kipf2016semi, zhuang2018dual, oono2019graph, hu2019hierarchical}. In practice, real-wold graphs grow rapidly and novel node classes could emerge incrementally in different time periods. For example, E-commerce platforms like Amazon can be naturally modeled by graphs, where products are represented as nodes, and the interactions between products (e.g., viewed by the same customers) are represented as edges. As newly emerged product categories will be continually added to the platform, the underlying GNN models need to handle a sequence of incremental learning sessions for product categorization, where each session introduces a set of novel classes. It is worth mentioning that, different from those base classes (in the first learning session) provided with abundant labeled data, only a few labeled samples are available for those newly emerged classes in any incremental learning session. Ideally, the desired GNN model is supposed to be capable of accurately recognizing those novel classes introduced in a new session while preserving the performance on all the ''seen'' node classes in previous sessions. Such \textit{Graph Few-shot Class-incremental Learning} (Graph FCL) problem has critical implications in both the academic and industrial communities. However, little effort has been devoted to this topic.

\label{challenge}The main challenges of the Graph FCL problem center around the so-called \emph{stability-plasticity} dilemma \cite{pulito1990two}, which is a trade-off between the preservation of previously learned graph knowledge and the capability of acquiring new knowledge. Specifically, the novel classes in each new learning session have much fewer nodes compared to the base classes, resulting in a severe class-imbalance problem \cite{hou2019learning, wu2019large, tao2020fewshot}. This may engender two potential problems: (1) on the one hand, if trained on all the data samples naively, the learned graph model could be substantially biased towards those base classes with significantly more nodes, resulting in the inertia to learn new node classes \cite{tao2020fewshot}. Moreover, to retain the existing knowledge, many of the methods from few-shot incremental learning \cite{ren2019incremental,yoon2020xtarnet,zhang2021fewshot} use a fixed feature encoder, which will not be updated after being pre-trained on base classes. Such a design will also exacerbate the difficulty of adapting the model to the new incremental learning tasks; (2) on the other hand, as the node classes from new tasks only have few-labeled samples, imposing the graph learning model to focus on new tasks will easily lead to overfitting to those new tasks and erase the existing knowledge for previously learned classes, which is known as \textit{Catastrophic Forgetting} \cite{kirkpatrick2017overcoming, kemker2018measuring, goodfellow2013empirical}. Considering the complex interactions between the nodes on graph-structured data, such learning errors would also be propagated on the graph and result in serious performance degradation. Thus, it is vital to explore and develop a new approach that can quickly adapt to the new class-incremental learning tasks while avoiding the forgetting of existing knowledge.

To address the aforementioned challenges of the studied problem, we first propose a new Graph Pseudo Incremental Learning (GPIL) training paradigm, which can facilitate the graph learning model to better adapt to new tasks. Initially, we split the original base dataset into base classes and pseudo novel classes with disjoint label spaces. Then, we pretrain the encoder on the base classes, and keep it trainable during the pseudo incremental learning process. For each episode during meta-training, all the few-shot node classification tasks are sampled from pseudo novel classes and base classes to mimic the incremental process in the evaluation phase. This way we obtain abundant meta-training episodes to learn a transferable model initialization for the incremental learning phase. We then propose a \textit{Hierarchical Attention Graph Meta-learning} framework, HAG-Meta, which can effectively handle the stability-plasticity dilemma. Specifically, our framework uses a dynamically scaled cross-entropy loss regularizer where the scale factors\cite{lin2017focal, menon2020long} are multiplied to each task-level loss to adjust their contribution for model training. Ideally, the scaling factors can help the model to down-weight the contribution of easy or insignificant tasks while focus on those hard or important tasks. Due to the fact that tasks in the Graph FCL problem have a hierarchical structure (nodes form classes, classes form tasks), we propose a hierarchical attention module that automatically captures the importance of different tasks and learns the scaling factors. On one hand, the \textit{Task-Level Attention} (TLA) will estimate the importance of each task based on their aggregated prototypes and output the scaling factors to balance the contribution of different tasks. On the other hand, the \textit{Node-Level Attention} (NLA) aims to learn prototypes that maintain a better balance between existing and novel knowledge within nodes and provide them to TLA. Being progressively trained in GPIL, the hierarchical attention module can gradually obtain the generalizability to produce the scaling factors for both encountered tasks and subsequent tasks. Training with this dynamically scaled regularizer, the proposed model will not only achieve better old knowledge consolidation but also acquire principled adaptability to new knowledge with merely limited data samples. The effectiveness of the proposed framework is validated with comprehensive experiments on three real-world datasets. The contribution of this work can be summarized as follows:
\begin{itemize}
  \item \textbf{\emph{Problem:}} We present a novel Graph Few-shot Incremental Learning problem and formulate it with node classification tasks, where the model is tasked to accomplish node classification on base classes and all few-shot novel classes encountered during incremental sessions.
  \item \textbf{\emph{Algorithm:}} We propose Hierarchical Graph Attention modules tailored for the Graph FCL problem, and design a Graph Pseudo Incremental Learning paradigm to enable effective training to mimic the environment in the evaluation phase.
  \item \textbf{\emph{Evaluation:}} Experiments on the Amazon-clothing, Reddit, and DBLP datasets show that the proposed framework significantly outperforms the baseline and other related state-of-the-art methods by a considerable margin.
\end{itemize}

\section{Related Work}

\subsection{Class-incremental Learning}
\textit{Incremental learning} (IL) \cite{ren2019incremental, yoon2020xtarnet}, also known as continual learning or lifelong learning, has drawn growing attention recently. IL aims to train machine learning models to acquire new knowledge while preserving the utmost existing knowledge. In this work, we mainly focus on \textit{Class-incremental Learning} (CIL) where novel classes emerge in subsequent sessions and the model is tasked to fulfill classification on all the classes it has encountered, rather than \textit{Task-incremental Learning} (TIL), where usually a task identifier is available, so the model can have multiple classifiers and finish the final classification on classes in a single task \cite{kim2021splitandbridge}. The mainstream of CIL methods can be categorized into two families. The first family includes replay-based methods \cite{lopez2017gradient, rebuffi2017icarl, rolnick2019experience, tang2021graphbased}, which maintain a subset of previous samples, and train models together with samples in the new
session. The other family of methods is regularization-based methods \cite{kirkpatrick2017overcoming, chaudhry2018riemannian}, where various regularizers are proposed to regularize the parameters of a neural network so that more important parameters concerning the previous task can be protected when models are trained on each new task. A common choice of regularizer is a \textit{Knowledge Distillation} (KD) \cite{hinton2015distilling} based loss proposed in LwF \cite{li2017learning}. iCaRL \cite{rebuffi2017icarl} is the first work that combines both replay and KD regularization methods, and puts forward the data imbalance problem between old classes and novel classes in CIL. A series of work focuses on this problem \cite{zhao2020maintaining, wu2019large}. \cite{ren2019incremental,yoon2020xtarnet,zhang2021fewshot} further extend the situation to the \textit{Few-shot Class-incremental Learning} (FCL) setting on image domain. For the FCL setting, KD performance will degrade tremendously due to the extreme scarcity of samples in novel classes. Instead, to overcome catastrophic forgetting, those methods usually adopt a decoupling method, where the encoder is fixed after pre-training, and extra modules are involved for learning incremental classes during meta-training. However, directly applying those FCL methods to the graph domain can lead to drastic performance degradation. Nodes in a graph are not i.i.d. data as usually assumed for images. Their representations are learned via sampling and aggregation from their neighbors. Fixing the encoder, if nodes in novel classes in an impending session are densely linked with nodes in base classes, their representation can have evident overlap, and the boundaries between those classes will be blurred. Since no existing work is suitable for FCL on graphs, our paper aims at bridging this gap.

\subsection{Graph Few-shot Learning}
Graph Neural Network (GNN) \cite{cao2016deep, defferrard2016convolutional, henaff2015deep, kipf2016semi, hamilton2018inductive, velivckovic2017graph, xu2018powerful} is a family of deep neural models tailored for graph-structured data, which has been widely used in various applications, such as recommendation~\cite{wang2020next}, anomaly detection~\cite{ding2019deep}, and text classification~\cite{ding2020more}. Generally, GNNs exploit a recurrent neighborhood aggregation strategy to preserve the graph structure information and transform the nodes' attributes simultaneously. For instance, variants of GCN \cite{cao2016deep, defferrard2016convolutional, henaff2015deep, kipf2016semi}, GraphSAGE \cite{hamilton2018inductive}, GAT \cite{velivckovic2017graph}, and GIN \cite{xu2018powerful} put forward different aggregation schemes to try to enhance the representation power of GNN. However, all those conventional GNNs may easily fail to learn expressive node representation when the labeled nodes are extremely scarce. Recently, increasing research attention has been devoted to graph few-shot learning problems. Especially, the episodic meta-learning paradigm \cite{thrun2012learning} has become the most popular strategy for this problem, which transfers knowledge learned from many similar FSL tasks. Based on it, Meta-GNN \cite{zhou2019meta} applies MAML \cite{finn2017modelagnostic} to tackle the low-resource learning problem on graph. Furthermore, RALE \cite{liu2021relative} uses GNNs to encode graph path-based hubs and capture the task-level dependency, to achieve knowledge transfer. GPN \cite{ding2020graph} adopts Prototypical Networks \cite{snell2017prototypical} to make the classification based on the distance between the node feature and the prototypes. AMM-GNN~\cite{wang2020graph} leverages an attribute-level attention mechanism to characterize the feature distribution differences between different tasks and learns more meaningful transferable knowledge across tasks. However, all those methods cannot be generalized to the Class-incremental learning scenarios, where the model is tested not only on the novel classes in the current task but also on all the classes in previous tasks. Catastrophic forgetting will erase the knowledge specific for previously learned classes.

\section{Problem Statement}
Formally, a graph $G = (\mathcal{V}, \mathcal{E}, \textbf{X})$, where $\mathcal{V}$, $\mathcal{E}$, and $\textbf{X}$ denote the set of nodes, edges and node features respectively, can be alternatively represented by $G = \{\textbf{A}, \textbf{X}\}$, where $\textbf{A}$ is the adjacency matrix. The Graph FCL task assumes the existence of a sequence of homogeneous datasets within a graph, $i.e.$,$\;\mathcal{D} = \{\mathcal{D}^{0}, \mathcal{D}^{1},..., \mathcal{D}^{i}..., \mathcal{D}^{T}\}$. In any session $i$,  $\mathcal{C}^i$, the label space of the dataset $\mathcal{D}^{i}$, has no overlapping with the label space of any other session, $i.e.$,$ \; \forall i, j \in \{0,...,T\}, i \neq j, \mathcal{C}^i \cap \mathcal{C}^j = \varnothing$. Then, the dataset in each learning session can be represented as $\mathcal{D}^i = \{\textbf{A}_{\mathcal{C}^{i}}, \textbf{X}_{\mathcal{C}^{i}}\}$, where $\textbf{A}_{\mathcal{C}^{i}}$ denotes the attributes of nodes whose labels belong to the label space $\mathcal{C}^{i}$. Notably, in the first session, the dataset $\mathcal{D}^0$ is a relatively large dataset where a sufficient amount of data is available for normal semi-supervised node classification training. The classes in $\mathcal{D}^0$ are the \textit{base classes}. Datasets in following sessions, $\mathcal{D}^i \in \mathcal{D}, i \neq 0$, are few-shot datasets, the classes in which are named as \textit{novel classes}. Now, we present the formal definition of a Few-shot Class-incremental Node Classification task:
\begin{definition}
  \textbf{Few-shot Class-incremental Node Classification on Graphs}: For a specific session $i$, given a graph $\mathcal{G} = (\textbf{A}, \textbf{X})$, and a set of support nodes with labels, $\mathcal{S}^i$, from the label space $\mathcal{C}^{i}$, the model is tasked to predict labels for the nodes in corresponding query set $\mathcal{Q}^i$. The label space of the query set $\mathcal{Q}^i$ includes the base set $\mathcal{C}^0$, all the novel sets in previous sessions $\{\mathcal{C}^1, \mathcal{C}^2,..., \mathcal{C}^{i-1}\}$, and the novel set encountered in the current session $\mathcal{C}^i$.
\end{definition}

For each session $i$, we have such a Few-shot Class-incremental Node Classification task $\mathcal{T}^{i}$. In the corresponding support set $\mathcal{S}^{i}$, we denote the number of novel classes as $N$ and the number of support nodes in each class as $K$. This task is named as an \textbf{$\bm{N}$-way $\bm{K}$-shot} incremental node classification task. Alternatively, learning through that sequence of datasets can be represented as a sequence of tasks: $i.e.$,$ \;T = \{\mathcal{T}^{0}, \mathcal{T}^{1},..., \mathcal{T}^{i}..., \mathcal{T}^{T}\}$. In essence, we want our graph model able to retain a decent performance when fulfilling the classification on both base and novel classes. 

\section{Methodology}
\begin{figure*}[h]
  \centering
  \includegraphics[width=\linewidth]{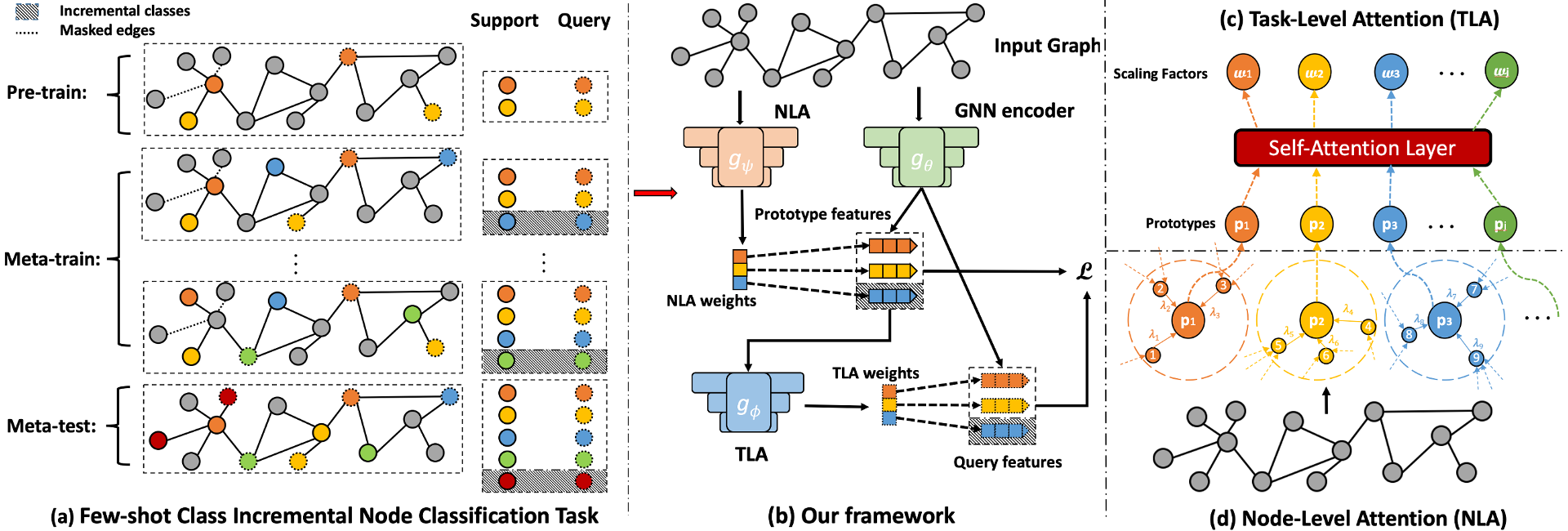}
  \caption{(a) Illustration of Few-shot Class-incremental Node Classification Task. (b) The illustration of our framework: HAG-Meta, as in Section \ref{model}. (c) Structure of the Task-Level Attention module, computed with self-attention layer, as in Section \ref{loss}. (d) Structure of the Node-Level Attention module. We adopt GCN layers to generate the weight,  as in Section \ref{nla}.}
  \Description{Data split strategy}
  \label{fig:frame}
\end{figure*}

In this section, we introduce a Hierarchical Attention Graph Meta-learning framework, \textbf{HAG-Meta} for solving Graph FCL problem. We first describe our proposed training procedures in Section \ref{procedure}. Then, we present the model proposed in Section \ref{model}. The overview of HAG-Meta is shown in Figure \ref{fig:frame}. Pseudocode-style algorithm descriptions are given in Algorithm \ref{alg:1} and Appendix \ref{app:IS}.

\subsection{Graph Pseudo Incremental Learning}
\label{procedure}

\textbf{Data Splits:} To solve the Graph FCL problem, We split the dataset $\mathcal{D}$ into two folds, $\mathcal{D}_{base}$ and $\mathcal{D}_{novel}$, with disjoint categories. $\mathcal{D}_{base}$ is randomly split into $\mathcal{D}_{base/tr}$, $\mathcal{D}_{base/val}$, and $\mathcal{D}_{base/test}$ for pre-training. $\mathcal{D}_{novel}$ has three splits of $\mathcal{D}_{novel/tr}$, $\mathcal{D}_{novel/val}$, $\mathcal{D}_{novel/test}$, with disjoint categories. To simulate the Graph FCL problem, the target few-shot data $\{\mathcal{D}^{1},..., \mathcal{D}^{i}..., \mathcal{D}^{T}\}$ are sampled from $\mathcal{D}_{novel/test}$, and the corresponding nodes and edges are masked during pre-training and meta-training. The base data $\mathcal{D}^{0}$ consists of $\mathcal{D}_{base}$ and $\mathcal{D}_{novel/tr}$, where our proposed Graph Pseudo Incremental Learning is conducted. Details are given next. 

\textbf{Pre-training:} We pre-train a GNN-based encoder $g_{\theta}$ on the split $\mathcal{D}_{base/tr}$, following the normal semi-supervised node classification process. The encoder is still trainable after pre-training.

\textbf{Meta-training:} To learn an initialization with more transferable meta-knowledge within the graph, here, we propose the \textbf{Graph Pseudo Incremental Learning (GPIL)} paradigm, where a model would be trained on $\mathcal{D}_{base/tr}$ and $\mathcal{D}_{novel/tr}$. Similar to the episodic meta-learning strategy, during each session $i$, we randomly sample $N$ novel categories from $\mathcal{D}_{novel/tr}$, and $K$ nodes per category to form a novel support set $\mathcal{S}^i_{novel} = \{(\textbf{x}^i_j, y^i_j)\}_{j=1}^{N \times K}$. The query set is composed of samples from the base categories, the novel categories from the previous sessions, and $N$ novel categories in the current session. We sample $K$ samples from all those categories to form a query set $\mathcal{Q}^i = \mathcal{Q}^i_{base} \cup \mathcal{Q}^i_{novel}$. The novel categories in each session will be merged into the base categories for the next session. We will cache those novel support nodes in $\mathcal{S}^i_{novel}$ for the classification of old classes in later sessions. So similarly, $\mathcal{S}^i = \mathcal{S}^i_{base} \cup \mathcal{S}^i_{novel}$. During each session $i$, the parameters are updated by the loss proposed in Section \ref{loss} for the classification of queries in $\mathcal{Q}^i$. During training, we reset the base categories and novel categories whenever the number of left novel categories is less than $N$. In this way, the proposed model can be trained on an arbitrary number of episodes despite the limitation of the number of novel categories in $\mathcal{D}_{novel/tr}$, which is crucial for training the attention models (See details in Section \ref{model}). 

\textbf{Evaluation:} For each evaluation session, we randomly sample $N$ novel categories from $\mathcal{D}_{novel/test}$ (or $\mathcal{D}_{novel/val}$ for validation). The proposed model is fine-tuned on the sample. Base samples for test are sampled from $\mathcal{D}_{base/test}$ (or $\mathcal{D}_{base/val}$ for validation) and $\mathcal{D}_{novel/train}$. For the next session, those novel categories are merged into base categories.

\subsection{Model}
\label{model}
The baseline model we deploy is based on Prototypical Network (PN)~\cite{snell2017prototypical}. We replace the multilayer perceptron (MLP) encoder in the original PN with a GNN encoder $g_{\theta}$. We call it Proto-GNN for convenience. Then, given a graph $\mathcal{G} = (\textbf{A}, \textbf{X})$, the latent features can be defined as:
\begin{equation}
    \textbf{Z} = g_{\theta}(\textbf{A}, \textbf{X})
\end{equation}
As mentioned in \cite{snell2017prototypical}, the vanilla way to compute a prototype for a category is the average over all the features of nodes in the category:
\begin{equation}
\label{PN}
    \textbf{p}_k = \frac{1}{|\mathcal{S}_k|} \sum_{j \in \mathcal{S}_k} \textbf{z}_{j}
\end{equation}
Where $\mathcal{S}_{k}$ is the set of all the support nodes index in category $k$ and $|\mathcal{S}_{k}|$ is its cardinality. Then, a distance function $d$ is used to generate the distribution over classes for a query node $v_q$ based on a softmax over distances to the prototypes in the latent space:
\begin{equation}
\label{clf}
    p(y = k | q) = \frac{\exp(-d(\textbf{z}_q, \textbf{p}_k))}{\sum_{k^\prime} \exp(-d(\textbf{z}_q, \textbf{p}_{k^\prime}))}
\end{equation}
For the choice of $d$, we use squared Euclidean distance, which has been shown as a simple and effective distance function \cite{snell2017prototypical}.

To overcome the class-imbalance challenge in Section \ref{challenge}, we here propose our Hierarchical-Attention based Graph Attention module. The goal is to learn a strong regularizer that can dynamically scale the contribution for nodes in different tasks. Tasks in the Graph FCL problem naturally have a hierarchical structure(nodes form classes, classes form tasks). Thus, we propose a two-level hierarchical attention mechanism: \textit{Task-Level Attention} and \textit{Node-Level Attention} to estimate the importance of different tasks.

\subsubsection{Task-Level Attention}
\label{loss}
To deal with the challenge that our model may overfit onto base or novel classes, we here propose the \textbf{Task-Level Attention} (\textbf{TLA}) to estimate the importance of classes learned in different tasks. Ultimately, we want to learn a series of scaling factors for the loss, which should be competent to automatically down-weight the contribution of easy or insignificant tasks during meta-training and rapidly focus the model on hard or important tasks. As described in the training procedure, during meta-training, the query nodes are from classes in all the previous training sessions. In this case, we find that a task-weighted loss will serve the purpose. For a PN, the prototypes matrix, $\textbf{P}^i = \{\textbf{p}_k^i\}$ ($\textbf{P}^i \in \mathbb{R}^{|\mathcal{S}_k| \times h}$, where $h$ is the size of prototype features), of the support nodes, $\mathcal{S}^i$, in a certain session $i$, serves as the classifier for the queries, $Q^i$. Hence, we can make the hypothesis that the prototypes in sessions are representative enough to express the knowledge of the task $\mathcal{T}^i$. Based on the self-attention by \cite{vaswani2017attention}, TLA aims at learning the attentions (scaling factors) $\textbf{w}^{i, j} \in \textbf{W}$, ($\forall j \in [1, i]$, $\textbf{W} \in \mathbb{R}^{1 \times i}$) between the current task, $\mathcal{T}^i$, and all the tasks the model has been trained on, $ \{\mathcal{T}^1, \mathcal{T}^2,...,\mathcal{T}^i\}$, including the current task. The desirable property of TLA is that the attention mechanism is inductive and permutation invariant, which suits the Graph FCL problem where novel tasks and classes come in sequence. The structure of the TLA model is shown in Figure \ref{fig:frame} (c). We use $g_\psi$ to denote the TLA generator. Because the number of classes in the base is much larger than that of each novel task, we first use MLPs to project the prototypes of all the tasks into the same size:
\begin{equation}
    \textbf{u}^j = MLP(\textbf{p}^j), \forall j \in [1, i]
\end{equation}
Where $\textbf{u}^j$ is the projections of the prototypes $\textbf{p}^j$ at session $j$. Then, the weights $\mathcal{W}$ can be computed as:
\begin{equation}
    \textbf{w}^{i,j} = \frac{\exp{(\textbf{u}^{i} \cdot \textbf{u}^j)}}{\sum_{j^{\prime}=1}^{i} \exp{(\textbf{u}^{i} \cdot \textbf{u}^{j^{\prime}})}}
\end{equation}
For each task, the weight $\mathcal{W}$ is determined by the number of classes in the task, and then normalized by the number of classes in that task (different classes in the same task share the same weight):
\begin{equation}
    \textbf{W}_C = \frac{\Tilde{\textbf{W}}}{|C^j|}
\end{equation}
where $|C^j|$ is the number of classes in session $j$, and $\Tilde{\textbf{W}} \in \mathbb{R}^{1 \times |C|}$ is the expanded weight vector of $\textbf{W}$. $|C|$ is the number of all the classes having been seen. And $\textbf{W}_C$ is the target scaling factor of classes within the task. With all the factors $\textbf{W}_C$ computed, finally we introduce our \textbf{TLA Loss}, which is the 
Cross-Entropy Loss (CEL) scaled by the TLA scaling factors $\textbf{W}_C$:
\begin{equation}
\label{TLA}
    \mathcal{L}_{TLA} = \sum_{k \in C} w_k \cdot [y_k \cdot \log(\hat{y_k}) + (1-y_k) \log(1-\hat{y_k})] 
\end{equation}
The final loss is:
\begin{equation}
\label{total}
\begin{split}
\mathcal{L} &= \mathcal{L}_{CEL} + \mathcal{L}_{TLA} \\
&= \sum_{k \in C} (1 + w_k) \cdot [y_k \cdot \log(\hat{y_k}) + (1-y_k) \log(1-\hat{y_k})]
\end{split}
\end{equation}
$\mathcal{L}_{CEL}$ will function as the initialization of task contribution weight, and $\mathcal{L}_{TLA}$ will turn into a regularizer, which adjusts the contribution according to the importance of tasks.

\subsubsection{Node-Level Attention}
\label{nla}
While TLA can weigh the importance of different graph tasks, it cannot fully capture the knowledge within the graph structure, which may lead to the inaccurate importance measured for tasks. To incorporate TLA with the graph knowledge, we propose to use a \textbf{Node-Level Attention (NLA)}, $\Lambda = \{\lambda_{j}\}, j \in \mathcal{S}^i$, to adjust the representation of the prototype features learned from the GNN encoder $g_{\theta}$ in each session $i$, where $\lambda_{j}$ is the attention weight of the novel support nodes $v_j$ to the prototype. We want the NLA to maintain the balance of existing knowledge and novel knowledge for each node. The schematic diagram of NLA can be found in Figure \ref{fig:frame} (d). We propose to use the Graph Convolutional Network (GCN) \cite{kipf2016semi} to calculate the final NLA. The propagation rule in the $l$th layer of our GCN can be represented as:
\begin{equation}
    \textbf{h}_j^l = \sigma' (\textbf{R}^{l}(\textbf{h}_{j}^{l-1} + \sum_{j^\prime \in \mathcal{N}_j} \textbf{h}_{j}^{l-1} / {\sqrt{d_j d_{j^\prime}}}))
\end{equation}
where, at the $l$th GCN layer, $\textbf{h}_j^l$ is the latent NLA representation of node $v_j$, $\sigma'$ is an activation function, $\textbf{R}^{l}$ is the learned weight matrix, $\mathcal{N}_j$ is a set of nodes adjacent to node $v_j$, $d_j$ and $d_{j^\prime}$ are the node degrees of node $v_j$ and node $v_{j^\prime}$ respectively. We set $\textbf{h}_j^0 = \textbf{X}_j$. Then, we use an MLP to project the latent NLA in the last ($L$th) GCN layer into scalar:
\begin{equation}
    \lambda_j = MLP(\textbf{h}_j^L)
\end{equation}
Next, we apply the centrality adjustment method proposed in \cite{page1999pagerank}:
\begin{equation}
    \Tilde{\lambda_j} = \sigma(log(degree(v_j)+\epsilon) \cdot \lambda_j)
\end{equation}
where $\sigma$ is the sigmoid function, $\epsilon$ is a small constant. Finally we use softmax function to normalize the NLA:
\begin{equation}
    \lambda_j = \frac{\exp(\Tilde{\lambda_j})}{\sum_{j^{\prime}\in \mathcal{S}^i} \exp(\lambda_{j^\prime})}
\end{equation}
Then, we modify the original strategy, Eq.\eqref{PN}, to calculate prototype in PN with NLA:
\begin{equation}
    \textbf{p}_k = \sum_{j \in \mathcal{S}_k} \lambda_j \textbf{z}_i
\end{equation}
With the adjusted prototypes, we can then use Eq.\eqref{clf} to get the final label. An overview of the incremental training procedure of each session is given in Appendix \ref{app:IS}.

\begin{algorithm}[t!]
\caption{\textsc{HAG-Meta}}
\label{alg:1}
\begin{algorithmic}[1]
\renewcommand{\algorithmicrequire}{\textbf{Input:}}
\renewcommand{\algorithmicensure}{\textbf{Output:}}
\REQUIRE Dataset $\mathcal{D}$, number of sessions $M$ for GPIL, number of target evaluation sessions $T$ for evaluation, random initialized GNN model $g_\theta$, random initialized TLA weighter $g_\phi$, and NLA weigter $g_\psi$.
\ENSURE Trained models: $g_\theta$, $g_\phi$, and $g_\psi$.

\COMMENT{\slash\slash \space \texttt{Data split}}
\STATE Split dataset $\mathcal{D}$ into $\mathcal{D}_{base/tr}$, $\mathcal{D}_{base/val}$, $\mathcal{D}_{base/test}$,\\ $\mathcal{D}_{novel/tr}$, $\mathcal{D}_{novel/val}$, and $\mathcal{D}_{novel/test}$.

\COMMENT{\slash\slash \space \texttt{GNN back bone Pre-training}}
\STATE Pre-train $g_\theta$ on $D_{base/tr}$ in normal supervised learning.

\COMMENT  {\slash\slash \space \texttt{Meta training (GPIL)}}
\WHILE {$i \le M$}
\STATE Sample a Graph FCL task from  $\mathcal{D}_{novel/tr}$ and $\mathcal{D}_{base/tr}$ according to Section \ref{procedure}: $\mathcal{T}^i_{tr} = \{\mathcal{S}^i, \mathcal{Q}^i\}$.
\STATE Do one incremental training session:

$\mathcal{A}^{i}$, $g_\theta$, $g_\phi$, $g_\psi$ = \textsc{IncrementalSession}($\mathcal{T}^{i}_{tr}$, $g_\theta$, $g_\phi$,  $g_\psi$)

\ENDWHILE

\COMMENT{\slash\slash \space \texttt{Evaluation}}

\WHILE {$j \le T$}
\STATE Sample a Graph FCL task from  $\mathcal{D}_{novel/test}$ and $\mathcal{D}_{base/test}$ according to Section \ref{procedure}: $\mathcal{T}^j_{test} = \{\mathcal{S}^j, \mathcal{Q}^j\}$.
\STATE Do one incremental training session:

$\mathcal{A}^{j}$, $g_\theta$, $g_\phi$, $g_\psi$ = \textsc{IncrementalSession}($\mathcal{T}^{j}_{test}$, $g_\theta$, $g_\phi$,  $g_\psi$)

\ENDWHILE
%\RETURN $P$
\end{algorithmic}
\end{algorithm}

\section{Experiments}
In this section, we present the evaluation of our framework: HAG-Meta. We first introduce the used datasets and compared methods. Then, we show the result and analysis of the comparative study. Furthermore, we conduct comprehensive ablation experiments to validate the effectiveness of individual components in the proposed framework and study their characteristics. Also, we compare the result of our model with that of the best baseline under different $N$-way $K$-shot settings. Finally, to illustrate the advantage of our model, we visualize the learned embeddings.
\subsection{Experiment Settings}
\textbf{Evaluation Datasets}. We conduct our experiments \footnote{Codes are avalable at \url{https://github.com/Zhen-Tan-dmml/GFCIL}} on three widely used graph Few-shot learning datasets: Amazon-Clothing \cite{mcauley2015inferring}, DBLP \cite{tang2008arnetminer}, and Reddit \cite{hamilton2018inductive}. More details about the datasets can be found in Appendix \ref{app:data}. The statistic are shown in Table \ref{tab:table1}.

\textbf{Compared Methods}. In this paper, we compare our HAG-Meta framework with the following methods:
\label{methods}
\begin{itemize}
  \item Prototypical Networks on graphs: As discussed in Section \ref{model}, we implement Proto-GNN with two different encoders: GCN \cite{kipf2016semi} and GAT \cite{velivckovic2017graph}, to reveal the inability of GNN to deal with the Graph FCL problem. We denote them as Proto-GCN and Proto-GAT.
  \item State-of-the-art Graph Few-shot learning methods: Meta-GNN \cite{zhou2019meta} and GPN \cite{ding2020graph}.
  \item Continual learning methods on graph: ER-GNN \cite{zhou2021overcoming}, and classic iCaRL \cite{rebuffi2017icarl} with the encoder substituted by a GNN. These models do not consider the Few-shot learning setting. 
  \item Few-shot Class-incremental Learning: CEC \cite{zhang2021fewshot}. It is one of the state-of-the-art methods but is primarily for the image domain, so we replace the encoder with a GNN encoder.
\end{itemize}

\subsection{Implementation Details}
\label{detail}
In Table \ref{tab:table1}, we list the specific data split strategy for each dataset, following Section \ref{procedure}. Here, $|\mathcal{C}_{base}|$ is the number of classes for pre-training, $|\mathcal{C}_{novel/tr}|$ the number of classes for meta-training, and $|\mathcal{C}_{novel/test}|$ the number of target few-shot classes for evaluation. E.g., for the Amazon-Clothing dataset, we pre-train our GNN encoder on 20 categories of nodes, which are viewed as base categories. Another 30 categories are used for meta-train for providing pseudo novel categories. For evaluation, the model will be fine-tuned on a sequence of tasks consisting of nodes in the remaining 27 target categories. We stop the encoder pre-training when its validation accuracy stops improving for more than 10 epochs. For model implementation, please refer to Appendix \ref{app:imp} for detail.

\begin{table}[]
\begin{tabular}{|cccc|}
\hline
\multicolumn{1}{|l}{} & Amazon Clothing & DBLP    & Reddit     \\ \hline
\# Nodes              & 24,919          & 40,672   & 232,965     \\
\# Edges              & 91,680          & 288,270 & 11,606,919 \\
\# Features            & 9,034           & 7,202     & 602       \\
\# Labels             & 77              & 137      & 41        \\ \hline
Pre-train             & 20              & 37      & 11         \\
Meta-train            & 30              & 50      & 10         \\
Evaluation             & 27              & 50      & 20         \\ \hline
\end{tabular}
\caption{\label{tab:table1}Statistics of the expermental datasets.}
\vspace{-1cm}
\end{table}

\textbf{Evaluation Protocol:} We evaluate the model after each
session with the test set $\mathcal{D}^{i}$ (sampled from $\mathcal{D}_{novel/test}$). To reduce fluctuation, all the accuracy (Acc.) scores reported are averaged over 10 random seeds. We also calculate a performance dropping rate (PD) that measures the absolute accuracy drops in the last session w.r.t. the accuracy before the first evaluation session, $i.e.$,$ \; PD = \mathcal{A}^0 - \mathcal{A}^T$ where $\mathcal{A}^0$ is the classification accuracy in the last Meta-train session and $\mathcal{A}^T$ is the accuracy in the last session. To make the result more explicit, we define a new term: Relative Performance Dropping rate (RPD), which is the PD normalized by the initial accuracy, $i.e.$,$ \; RPD = \frac{PD}{\mathcal{A}^0}$.

\subsection{Comparative Results}
In this section, we present the comparison between our framework and the other four categories of baseline methods described in Section \ref{methods}. To the best of our knowledge, we are the first to investigate the Graph Few-shot Class-incremental Learning Problem. To fairly compare those methods, all methods except the basic GAT model share the same pre-trained GNN encoder as the proposed framework HAG-Meta, a 2-layer GCN. Also, when experimented on each dataset, they share the same random seeds for data split, leading to identical evaluation data.
We justify the advantage of the proposed framework from the following aspects:

\textbf{Performance Degradation} in Graph Few-shot Learning (GFSL) methods: A general observation is that, for those GFSL methods, their accuracy decreases substantially as new sessions emerge, especially for the first several sessions. 
Even though through meta-training, they have gained generalizability to a certain extent, the performance degrades constantly as the number of classes involved increases. This implies that the existing GFSL methods cannot maintain discriminative boundaries between all the base classes and novel classes. Without consolidated knowledge of classes in base and early sessions, traditional GFSL methods suffer grievously from Catastrophic Forgetting, as it adapts to classes in the latest episode.

\textbf{Limitation} in existing Incremental Learning methods. ER-GNN is one of the pioneers to task graph neural network models with a sequence of tasks. It adopts several Experience Replay methods to try to consolidate existing knowledge. However, the accuracy of ER-GNN diminishes tremendously when it is applied to a few-shot setting. Besides, iCaRL combines a piece of memory and knowledge distillation to consolidate existing knowledge. However, the limited number of samples in novel classes will affect the knowledge distillation process, leading to its overfitting onto old classes. Furthermore, the CEC method is one of the state-of-the-art few-shot Class-incremental learning methods for the image domain. It assumes that the data is i.i.d. distributed, which makes it overlook the topological relationships among tasks and nodes. It adopts a decoupling strategy, where the encoder is fixed after pre-training to retain existing knowledge. So when it is applied to graphs, the representation of nodes in base classes is fixed through all following sessions. But in graphs, the representation of novel classes nodes tightly depends on their neighboring nodes, which might belong to the base classes. This leads to indiscriminative boundaries between the base and the novel classes and unsatisfactory accuracy.

\textbf{Advantages} of the proposed HAG-Meta: Generally, HAG-Meta outperforms all baseline methods by a large margin, in terms of accuracy, PD, and RPD. Compared to the GFSL methods, the proposed penalty term $\mathcal{L}_{TLA}$ can effectively prevent catastrophic forgetting by regularization. In contrast with ER-GNN and iCaRL, our GPIL paradigm provides sufficient episodes to train the model such that it can learn an appropriate initialization to adapt to few-shot novel data. Also, we discard the decoupling strategy of the mainstream incremental learning methods on the image domain, like CEC. In contrast, we add node level attention, NLA, to learn the task and class dependency within the graph topological structure. The results in Table \ref{tab:table2} verify that HAG-Meta is robust against increasing numbers of sessions with novel classes in the Graph FCL problem.

\begin{table*}[]

\scalebox{1}{
\begin{tabular}{|cccccccccccccc|}
\hline
\multicolumn{14}{|c|}{\textbf{Amazon-Clothing dataset (3-way 5-shot)}}                                                                                                                                                                                                                                                           \\ \hline
\multirow{2}{*}{Method} & \multicolumn{10}{c}{Acc. in each session ($\%$) $\uparrow$}                                                                                                                          & \multirow{2}{*}{PD$\downarrow$} & \multirow{2}{*}{RPD$\downarrow$} & \multirow{2}{*}{\begin{tabular}[c]{@{}c@{}}improvement\\ (PD/RPD)\end{tabular}} \\ \cline{2-11}
                        & \textbf{0}     & \textbf{1}     & \textbf{2}     & \textbf{3}     & \textbf{4}     & \textbf{5}     & \textbf{6}     & \textbf{7}     & \textbf{8}     & \textbf{9}     &                     &                      &                                                                                 \\ \hline
Proto-GAT               & 60.22          & 44.32          & 41.87          & 38.28          & 35.15          & 32.25          & 30.67          & 28.54          & 26.54          & 25.43          & 34.79               & 57.77                & \textbf{(+1.43/+18.13)}                                                         \\
Proto-GCN               & 60.52          & 43.11          & 42.41          & 39.80          & 36.91          & 33.84          & 31.47          & 29.67          & 27.18          & 25.76          & 34.76               & 57.44                & \textbf{(+1.40/+17.80)}                                                         \\
Meta-GNN                & 79.62          & 65.29          & 63.46          & 58.49          & 56.36          & 54.13          & 51.76          & 49.50          & 46.32          & 43.07          & 36.55               & 45.90                & \textbf{(+3.19/+6.26)}                                                          \\
GPN                     & 80.76          & 66.98          & 64.46          & 62.63          & 60.03          & 55.48          & 52.35          & 50.76          & 48.15          & 45.83          & 34.93               & 43.25                & \textbf{(+1.57/+3.61)}                                                          \\ \hline
ER-GNN                  & 81.37          & 73.62          & 70.84          & 63.63          & 61.86          & 58.24          & 54.29          & 51.58          & 48.86          & 46.26          & 35.11               & 43.14                & \textbf{(+1.75/3.50)}                                                           \\
iCaRL                   & 79.43          & 73.92          & 66.87          & 63.19          & 60.28          & 56.04          & 53.48          & 50.33          & 47.75          & 45.82          & 33.61               & 42.31                & \textbf{(+0.25/+2.67)}                                                          \\
CEC                     & 82.05          & 73.28          & 70.46          & 64.85          & 62.19          & 60.29          & 54.86          & 52.27          & 50.69          & 48.22          & 34.28               & 41.77                & \textbf{(+0.92/+2.13)}                                                          \\ \hline
\textbf{Ours}           & \textbf{84.15} & \textbf{75.32} & \textbf{71.35} & \textbf{67.32} & \textbf{64.03} & \textbf{61.42} & \textbf{56.23} & \textbf{54.63} & \textbf{52.65} & \textbf{50.79} & \textbf{33.36}      & \textbf{39.64}       &                                                                                 \\ \hline
\end{tabular}}

\smallskip
\scalebox{1}{
\begin{tabular}{|cccccccccccccc|}
\hline

\multicolumn{14}{|c|}{\textbf{DBLP dataset (5-way 5-shot)}}                                                                                                                                                                                                                                                                      \\ \hline
\multirow{2}{*}{Method} & \multicolumn{10}{c}{Acc. in each session ($\%$) $\uparrow$}                                                                                                                          & \multirow{2}{*}{PD$\downarrow$} & \multirow{2}{*}{RPD$\downarrow$} & \multirow{2}{*}{\begin{tabular}[c]{@{}c@{}}improvement\\ (PD/RPD)\end{tabular}} \\ \cline{2-11}
                        & \textbf{0}     & \textbf{1}     & \textbf{2}     & \textbf{3}     & \textbf{4}     & \textbf{5}     & \textbf{6}     & \textbf{7}     & \textbf{8}     & \textbf{9}     &                     &                      &                                                                                 \\ \hline
Proto-GAT               & 38.75          & 32.04          & 23.64          & 19.65          & 19.43          & 18.64          & 18.86          & 18.59          & 18.35          & 18.23          & 20.52               & 52.96                & \textbf{(+3.41/+21.62)}                                                         \\
Proto-GCN               & 39.02          & 33.11          & 24.27          & 20.14          & 20.03          & 19.12          & 18.76          & 18.32          & 18.50          & 18.66          & 20.36               & 52.17                & \textbf{(+3.25/+20.83)}                                                         \\
Meta-GNN                & 40.75          & 34.28          & 33.68          & 32.18          & 30.45          & 28.45          & 24.34          & 23.18          & 23.07          & 22.23          & 18.52               & 45.44                & \textbf{(+1.41/+14.10)}                                                         \\
GPN                     & 41.39          & 34.53          & 32.18          & 32.00          & 31.25          & 30.67          & 29.16          & 27.64          & 25.34          & 23.13          & 18.26               & 44.12                & \textbf{(+1.15/12.78)}                                                          \\ \hline
ER-GNN                  & 46.74          & 41.70          & 38.55          & 36.82          & 35.03          & 34.28          & 33.32          & 31.59          & 30.67          & 28.86          & 17.88               & 38.25                & \textbf{(+0.77/6.91)}                                                           \\
iCaRL                   & 45.76          & 40.03          & 37.92          & 36.56          & 34.81          & 33.22          & 32.10          & 30.57          & 29.34          & 28.24          & 17.52               & 38.29                & \textbf{(+0.41/+6.95)}                                                          \\
CEC                     & 46.45          & 40.25          & 38.28          & 36.67          & 35.26          & 34.48          & 32.24          & 31.68          & 30.84          & 28.76          & 17.69               & 38.08                & \textbf{(+0.58/+6.74)}                                                          \\ \hline

\textbf{Ours}           & \textbf{54.59} & \textbf{47.89} & \textbf{46.45} & \textbf{45.05} & \textbf{43.83} & \textbf{41.38} & \textbf{39.67} & \textbf{39.07} & \textbf{38.15} & \textbf{37.48} & \textbf{17.11}      & \textbf{31.34}       &                                                                                 \\ \hline
\end{tabular}}

\smallskip
\scalebox{1}{
\begin{tabular}{|cccccccccccccc|}
\hline
\multicolumn{14}{|c|}{\textbf{Reddit dataset (2-way 3-shot)}}                                                                                                                                                                                                                                                                    \\ \hline
\multirow{2}{*}{Method} & \multicolumn{10}{c}{Acc. in each session ($\%$) $\uparrow$}                                                                                                                          & \multirow{2}{*}{PD$\downarrow$} & \multirow{2}{*}{RPD$\downarrow$} & \multirow{2}{*}{\begin{tabular}[c]{@{}c@{}}improvement\\ (PD/RPD)\end{tabular}} \\ \cline{2-11}
                        & \textbf{0}     & \textbf{1}     & \textbf{2}     & \textbf{3}     & \textbf{4}     & \textbf{5}     & \textbf{6}     & \textbf{7}     & \textbf{8}     & \textbf{9}     &                     &                      &                                                                                 \\ \hline
Proto-GAT               & 48.23          & 42.07          & 37.52          & 33.43          & 32.38          & 31.05          & 28.32          & 26.42          & 24.39          & 22.47          & 25.76               & 53.41                & \textbf{(+5.14/+19.43)}                                                         \\
Proto-GCN               & 48.04          & 42.77          & 37.35          & 35.04          & 33.62          & 31.21          & 28.04          & 25.86          & 24.42          & 22.61          & 25.43               & 52.94                & \textbf{(+4.81/+18.96)}                                                         \\
Meta-GNN                & 53.14          & 48.56          & 45.63          & 42.52          & 40.42          & 38.20          & 35.12          & 32.21          & 30.68          & 29.45          & 23.69               & 44.58                & \textbf{(+5.07/+10.60)}                                                         \\
GPN                     & 55.28          & 51.84          & 46.36          & 43.71          & 41.18          & 39.07          & 37.82          & 35.04          & 32.48          & 31.65          & 23.63               & 42.75                & \textbf{(+3.01/+8.77)}                                                          \\ \hline
ER-GNN                  & 52.86          & 47.29          & 45.28          & 43.56          & 41.08          & 40.02          & 38.42          & 36.77          & 33.26          & 30.47          & 22.39               & 42.36                & \textbf{(+1.77/+8.38)}                                                          \\
iCaRL                   & 54.62          & 50.58          & 48.72          & 46.23          & 44.84          & 42.16          & 40.29          & 38.65          & 35.74          & 33.28          & 21.34               & 39.07                & \textbf{(+0.72/+5.09)}                                                          \\
CEC                     & 57.68          & 53.13          & 50.63          & 48.37          & 46.76          & 43.13          & 41.28          & 39.68          & 37.42          & 36.61          & 21.07               & 36.53                & \textbf{(+0.45/+2.55)}                                                          \\ \hline
\textbf{Ours}           & \textbf{60.68} & \textbf{53.26} & \textbf{52.68} & \textbf{50.82} & \textbf{49.37} & \textbf{47.25} & \textbf{45.86} & \textbf{43.16} & \textbf{42.28} & \textbf{40.06} & \textbf{20.62}      & \textbf{33.98}       & \textbf{}                                                                       \\ \hline
\end{tabular}}
\caption{\label{tab:table2}Comparative Results on the three datasets under different N-way K-shot settings.}
\end{table*}

\subsection{Ablation Study}
In this section, we conduct more experiments to investigate the effectiveness of different components in our framework. We present the results of experiments on Amazon-Clothing datasets, under the 3-way 5-shot setting (similar results can be observed on the other datasets and settings). The results are shown in Table \ref{tab:table3}. For each method, the models share the same data splits for evaluation.

Specifically, the variant Proto-GNN stands for the Prototypical Graph Neural Networks baseline described in Section \ref{model}. NLA and TLA represent Node-Level Attention and Task-level Attention, respectively. A method without the TLA means the loss is the vanilla cross-entropy loss, without the $\mathcal{L}_{TLA}$ regularizer, see Eq.\eqref{total}. A method without the NLA means there is no prototype representation adjustment. All the prototype representations are the vanilla average of the encoder output, see Eq.\eqref{PN}. For methods without GPIL, the model is directly fine-tuned on datasets with target novel classes in different sessions.

\begin{table*}[t]
\scalebox{1}{
\begin{tabular}{|ccccccccccccc|}
\hline
\multirow{2}{*}{Method}               & \multicolumn{10}{c}{Acc. in each session ($\%$) $\uparrow$}                                                                                                                          & \multirow{2}{*}{PD$\downarrow$} & \multirow{2}{*}{RPD$\downarrow$} \\ \cline{2-11}
                                      & 0              & 1              & 2              & 3              & 4              & 5              & 6              & 7              & 8              & 9              &                     &                      \\ \hline
Proto-GNN                             & 60.52          & 43.11          & 42.41          & 39.80          & 36.91          & 33.84          & 31.47          & 29.67          & 27.18          & 25.76          & 34.76               & 57.44                \\ \hline
Proto-GNN + GPIL                      & 78.24          & 68.77          & 62.48          & 55.20          & 51.48          & 45.65          & 42.89          & 37.42          & 34.62          & 32.52          & 45.72               & 58.44                \\ \hline
Proto-GNN + NLA                       & 76.68          & 65.33          & 62.04          & 60.64          & 56.88          & 53.37          & 50.65          & 46.96          & 43.26          & 40.40          & 36.28               & 47.31                \\ \hline
Proto-GNN + TLA                       & 75.54          & 63.87          & 60.78          & 58.27          & 56.24          & 53.41          & 51.78          & 48.61          & 44.18          & 41.35          & 34.19               & 45.26                \\ \hline
Proto-GNN + NLA + GPIL                & 80.75          & 74.16          & 70.48          & 64.88          & 61.56          & 57.71          & 54.65          & 50.25          & 48.89          & 44.5           & 36.25               & 44.89                \\ \hline
Proto-GNN + TLA + GPIL                & 79.94          & 73.67          & 70.01          & 65.21          & 62.33          & 58.24          & 54.14          & 51.33          & 48.65          & 45.32          & 34.62               & 43.31                \\ \hline
Proto-GNN + NLA + TLA                 & 77.03          & 72.24          & 68.75          & 63.02          & 60.38          & 56.41          & 53.96          & 49.23          & 45.48          & 42.93          & 34.10               & 44.27                \\ \hline
\textbf{Proto-GNN + NLA + TLA + GPIL} & \textbf{84.15} & \textbf{75.32} & \textbf{71.35} & \textbf{67.32} & \textbf{64.03} & \textbf{61.42} & \textbf{56.23} & \textbf{54.63} & \textbf{52.65} & \textbf{50.79} & \textbf{33.36}      & \textbf{39.64}       \\ \hline
\end{tabular}}
\caption{\label{tab:table3}Ablation results on Amazon-Clothing dataset (3-way 5-shot).}
\vspace{-0.65cm}
\end{table*}

\begin{figure*}[ht!]
  \centering
  \includegraphics[width=\linewidth]{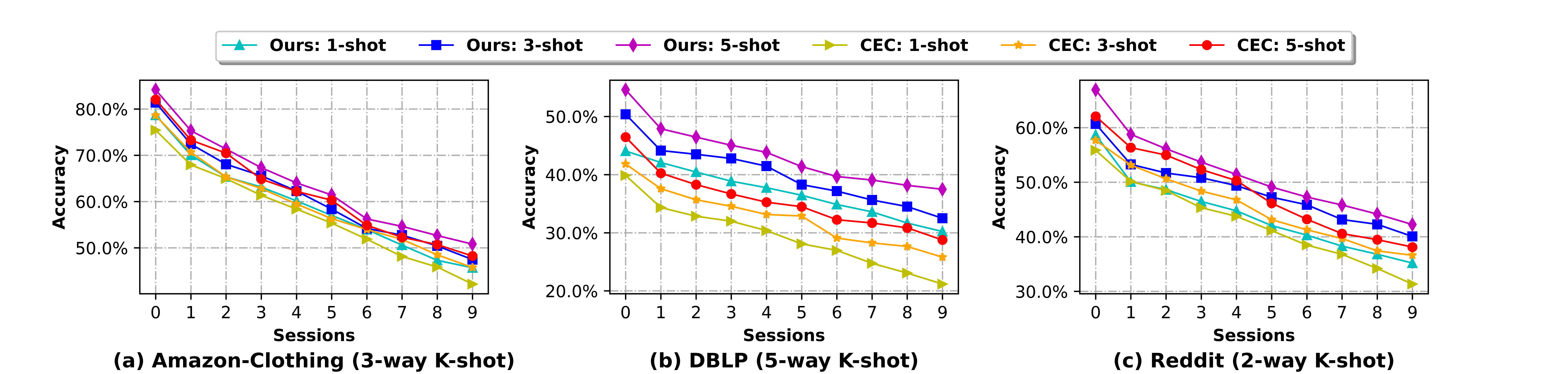}
  \caption{Parameter analysis on three real-world datasets: (a) Amazon-Clothing (b) DBLP (c) Reddit. For each dataset we experiment under three N-way k-shot settings.}
  \Description{Comparative result}
  \label{fig:para}
\end{figure*}

According to the results shown in Table \ref{tab:table3}, we can observe that each component in our proposed framework, HAG-Meta, is effective when tackling the Graph FCL problem:
\begin{itemize}
   \item GPIL: Comparison between methods with and without GPIL shows that GPIL can improve the accuracy. With more episodes of training, GPIL equips the hierarchical-attention module with better initialization to capture the importance among nodes and tasks.
  \item TLA and NLA: Generally, the scores show that both attention components help improve accuracy. Specifically, comparing methods with solely NLA or TLA, we can find that, at first several sessions, methods with NLA have higher scores, but their scores decrease at a sharper rate compared to methods with only TLA. This shows that NLA is better in terms of capturing the latent information within the graph data from limited novel class samples, but it suffers from catastrophic forgetting. In contrast, methods with TLA can maintain a higher score at later sessions, which echoes the purpose of our design, that the TLA loss can alleviate the forgetting problem by focusing more on important hard-to-learn tasks than insignificant easy-to-learn tasks.
  \item Our framework, HAG-Meta, containing NLA, TLA, and GPIL, as shown in the last line in Table \ref{tab:table3}, achieves the best performance. Through effective training in GPIL, the hierarchical attention components, TLA and NLA, have learned an initialization that can capture the topological information within the input graph to adapt to few-shot novel classes and mitigate forgetting of the existing knowledge simultaneously.
\end{itemize}

\subsection{Parameter Analysis}
In this section, we present extensive experiments to analyze the sensitivity of our HAG-Meta to the number of node classes ($N$-way) and support(query) set size ($K$-shot). For better comparison, we include the accuracy for both our HAG-Meta and the best comparable model CEC under different N-way K-shot settings for all three real-world datasets. As the result shown in Figure \ref{fig:para}, our model outperforms CEC in every setting we test. Plus, with more supervisory signals, the classification accuracies of both models are higher with larger support and query set.

\subsection{Visualization}
\begin{figure}[h]
  \centering
  \includegraphics[width=\linewidth]{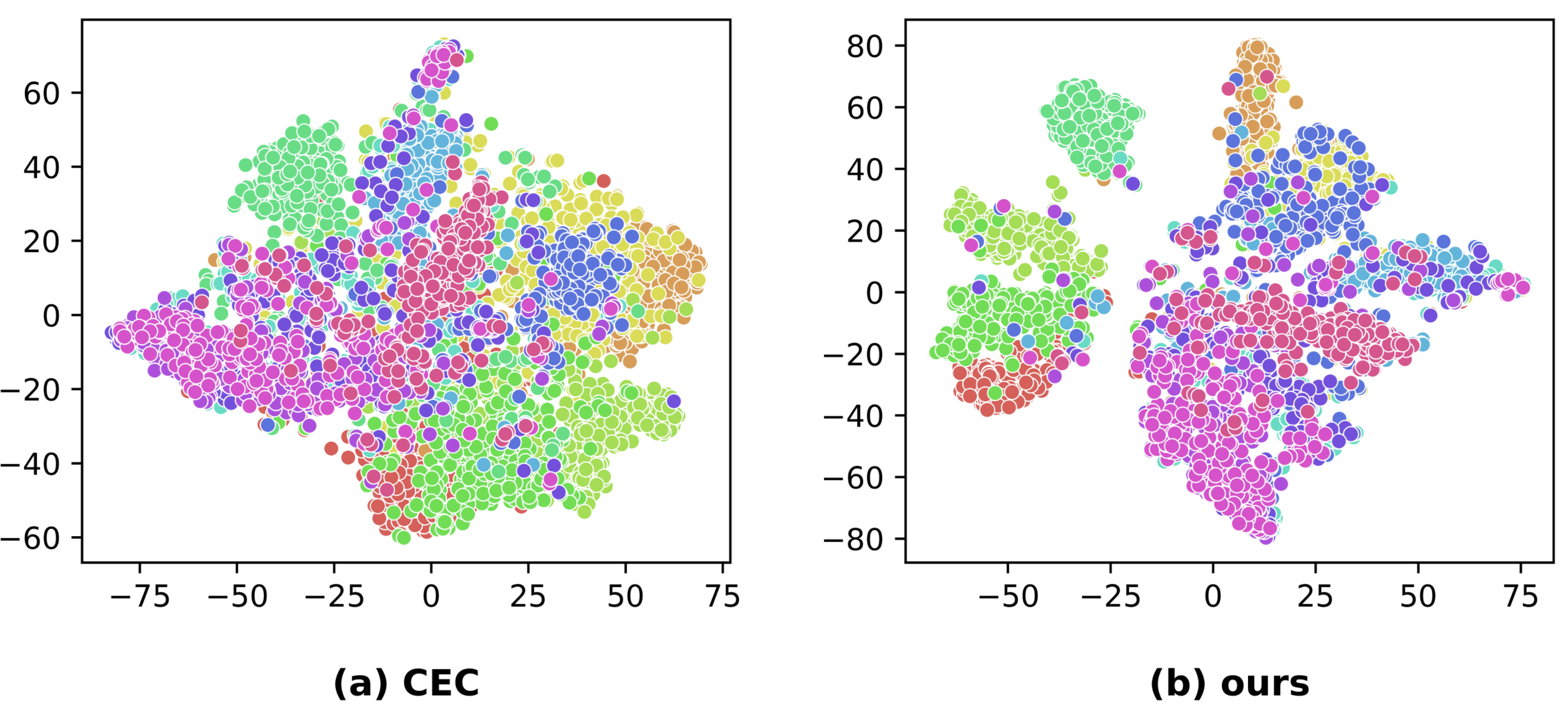}
  \caption{t-SNE embedding visualization: (a) CEC (b) ours.}
  \Description{Visualization}
  \label{fig:visual}
  \vspace{-0.3cm}
\end{figure}

To illustrate the effectiveness of our framework, we use the t-SNE \cite{van2008visualizing} method to visualize the embedding after the second session of evaluation for the large-scale Reddit dataset under the 2-way 5-shot setting. As shown in Figure \ref{fig:visual}, each color signifies a class. It's evident that our model is capable to produce projection embedding that elicits much more discriminative decision boundaries than our best baseline model CEC.

\section{Conclusion}
In this work, we propose a new problem, the Graph Few-shot Class-incremental Learning (Graph FCL) problem, aiming at obtaining a graph model that can adapt to new tasks with a restricted number of labeled training samples in novel classes, and simultaneously keep a good performance on old tasks. We formalize it with the node classification task and propose a novel framework called HAG-Meta. We present a novel Graph Pseudo Incremental Learning paradigm, which allows our model to learn a  generalizable initialization for the evaluation phase. Then we propose a hierarchical-attention-based module to solve the class-imbalance problem in Graph FCL. Primarily, the Task-Level Attention will be trained to estimate the importance between different tasks for backpropagation, while the Node-Level Attention incorporates the Task-Level Attention with the ability to capture the knowledge within the graph structure. A dynamically scaled loss regularizer is then computed from the task importance and automatically adjusts the contribution of different tasks for training. We conduct experiments on real-world datasets to demonstrate the effectiveness and advantage of our framework. 

Despite the promising results, the Graph FCL problem is far from being solved. In particular, many other graph settings are worth considering, such as the dynamic graph scenario, where the graph structure is continually evolving in different learning sessions. Besides, more sophisticated methods to design the regularizer, like from the perspective of causality, are worth considering. Also, we plan to study the scenario where the model cannot explicitly store any of the training data \cite{rebuffi2017icarl}, e.g. for privacy issues. A potential direction could be to investigate the efficient methods to store or reproduce embedded features \cite{kipf2016variational, Rannen_2017}.

% \section{Appendices}

\bibliographystyle{ACM-Reference-Format}
\bibliography{gfcil}

\pagebreak
\appendix
\section{Datasets Description}
\label{app:data}
In this section, we provide a detailed description of all those three datasets we experiment on. We follow the pre-processing procedures in \cite{ding2020graph}.
\begin{itemize}
  \item Amazon-Clothing \cite{mcauley2015inferring} is a product network built with the products in “Clothing, Shoes, and Jewelry” on Amazon. In this dataset, each product is considered as a node and its description is used to construct the node attributes. We use the substitutable relationship (“also viewed”) to create links between products. 
  \item DBLP \cite{tang2008arnetminer} is a citation network between academic papers where each node represents a paper, and the links are the citation relations among different papers. The paper abstracts are used to construct node attributes. The class label of a node is defined as the paper venue.
  \item Reddit \cite{hamilton2018inductive} is a large-scale post-to-post graph constructed with data sampled from Reddit, within which posts are represented by nodes and two posts are connected if they are commented by the same user. Each post is labeled with a community ID. 
\end{itemize}

\section{Implementation Detail}
\label{app:imp}
For the model implementation, we implement the proposed framework in PyTorch. Specifically, the graph encoder $g_\theta$ consists of two GCN layers \cite{kipf2016semi} with dimension size 32 and 16, respectively. Both of them are activated with the ReLU function. Regarding the TLA, We use a 3-layer MLP to map the base prototypes into the same size as the novel prototypes in each session, namely, $N$. For the NLA, it consists of one fully connected layer and two GCN aggregation layers \cite{kipf2016semi}. For each aggregation layer, we use ReLU function as the activation function. 

The framework is trained with Adam optimizers whose learning rates are set to be 0.005 initially with a weight decay of 0.0005. And the coefficients for computing running averages of gradient and square are set to be $\beta_1$ = 0.9, $\beta_2$ = 0.999. For each dataset, we meta-train the model over 1000 episodes with an early-stopping strategy.

\section{Pseudo-code Style Description of Each Incremental Session}
\label{app:IS}
\begin{algorithm}[]
\caption{\textsc{IncrementalSession}}
\begin{algorithmic}[1]
\renewcommand{\algorithmicrequire}{\textbf{Input:}}
\renewcommand{\algorithmicensure}{\textbf{Output:}}
\REQUIRE Sampled task $\mathcal{T} = \{\mathcal{S}, \mathcal{Q}\}$, $g_\theta$, $g_\phi$, and $g_\psi$.
\ENSURE Accuracy $\mathcal{A}$, trained models: $g_\theta$, $g_\phi$, and $g_\psi$.

\COMMENT{\slash\slash \space \texttt{One session of Graph FCL}}

\STATE Compute representation of nodes in $\mathcal{S}$ and $\mathcal{Q}$ with $g_\theta$.
\STATE Compute $\Lambda$ and $\textbf{p}$, with $g_\psi$.
\STATE Compute $\mathcal{W}_C$, and labels for nodes in $\mathcal{Q}^i$ with $g_\phi$.
\STATE Compute TLA Loss and total Loss, $\mathcal{L}_{TLA}$ and $\mathcal{L}$, with Eq.\eqref{TLA} and Eq.\eqref{total} for backprop.
\STATE Compute labels for nodes in $\mathcal{Q}$ and corresponding accuracy $\mathcal{A}$.

%\RETURN $P$
\end{algorithmic}
\end{algorithm}



\end{document}